\documentclass[]{article}

\usepackage{subfig}
\usepackage{placeins}
\usepackage{amsmath}
\usepackage{xcolor}
\usepackage{hyperref}
\usepackage{subfig}
\usepackage{graphicx}
\usepackage[round, authoryear]{natbib}
\usepackage{authblk}
\usepackage{fullpage}

\title{PyMatting: A Python Library for Alpha Matting}

\author[1]{Thomas Germer}
\author[1]{Tobias Uelwer}
\author[1]{Stefan Conrad}
\author[1]{Stefan Harmeling}
\affil[1]{Department of Computer Science, Heinrich-Heine-Universit\"at D\"usseldorf, Germany}
\affil[ ]{\normalfont\texttt{\{thomas.germer, tobias.uelwer, stefan.conrad, harmeling\}@hhu.de}}
\date{}
\begin{document}

\maketitle

\begin{abstract}
	An important step of many image editing tasks is to extract specific objects from an image in order to place them in a scene of a movie or compose them onto another background. 
	Alpha matting describes the problem of separating the objects in the foreground from the background of an image given only a rough sketch. 
	We introduce the PyMatting package for Python which implements various approaches to solve the alpha matting problem.
	Our toolbox is also able to extract the foreground of an image given the alpha matte. 
	The implementation aims to be computationally efficient and easy to use. 
	The source code of PyMatting is available under an open-source license at \url{https://github.com/pymatting/pymatting}.\\
\end{abstract}

\section{Introduction}
For an image $I$ with foreground pixels $F$ and background pixels $B$, alpha matting asks to determine opacities $\alpha$, such that the equality
\begin{equation}
I_i = \alpha_i F_i + (1-\alpha_i)B_i
\end{equation}
holds for every pixel $i$. This problem is ill-posed since, for each pixel, we have three equations (one for each color channel) with seven unknown variables.
The implemented methods rely on a trimap, which is a rough classification of the input image into foreground, background and unknown pixels, to further constrain the problem.
Giving the alpha matte, foreground estimation aims to extract the foreground $F$ from image $I$.

\section{Implemented Methods for Alpha Estimation}

\paragraph{Closed-form Matting:} 
\cite{levin2007closed} show that assuming local smoothness of pixel colors yields a closed-form solution to the alpha matting problem. 

\paragraph{KNN Matting:} 
\cite{lee2011nonlocal} and \cite{chen2013knn} use nearest neighbor information to derive closed-form solutions to the alpha matting problem which they note to perform particularly well on sparse trimaps. 

\paragraph{Large Kernel Matting:} 
\cite{he2010fast} propose an efficient algorithm based on a large kernel matting Laplacian.
They show that the computational complexity of their method is independent of the kernel size.

\paragraph{Random Walk Matting:} 
\cite{grady2005random} use random walks on the pixels to estimate alpha. 
The calculated alpha of a pixel is the probability that a random walk starting from that pixel will reach a foreground pixel before encountering a background pixel.

\paragraph{Learning Based Digital Matting:} 
\cite{zheng2009learning} estimate alpha using local semi-supervised learning. 
They assume that the alpha value of a pixel can be learned by a linear combination of the neighboring pixels.

\section{Implemented Methods for Foreground Estimation}

\paragraph{Closed-form Foreground Estimation:} For given $\alpha$, the foreground pixels $F$ can be determined by making additional smoothness assumptions on $F$ and $B$. 
Our toolbox implements the foreground estimation by \cite{levin2007closed}.

\paragraph{Multi-level Foreground Estimation:} 
Furthermore, the PyMatting toolbox implements a novel multi-level approach for foreground estimation.
For this method our toolbox also provides GPU implementations for OpenCL and CUDA.

\section{Installation and Code Example}

\begin{figure}
	\centering
	\subfloat[Input $I$]{{\fbox{\includegraphics[width=3cm]{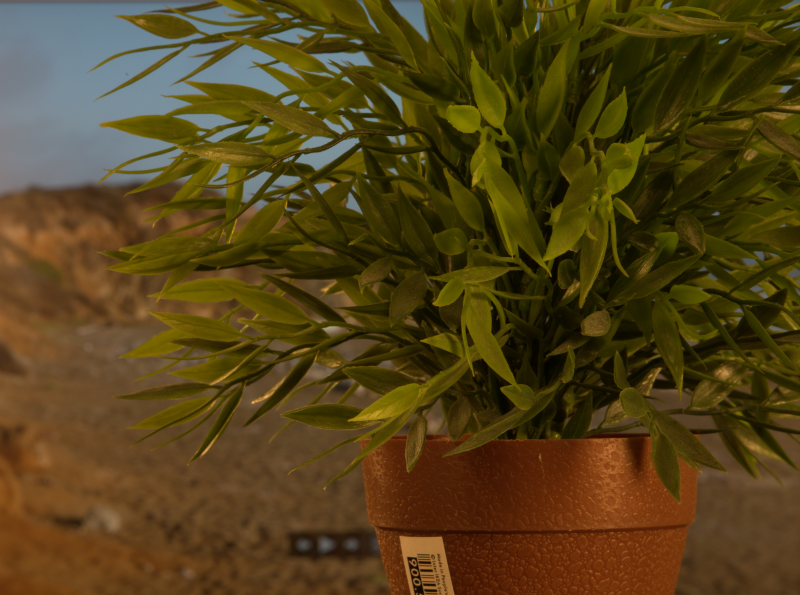}}}}
	\qquad
	\subfloat[Input trimap]{{\fbox{\includegraphics[width=3cm]{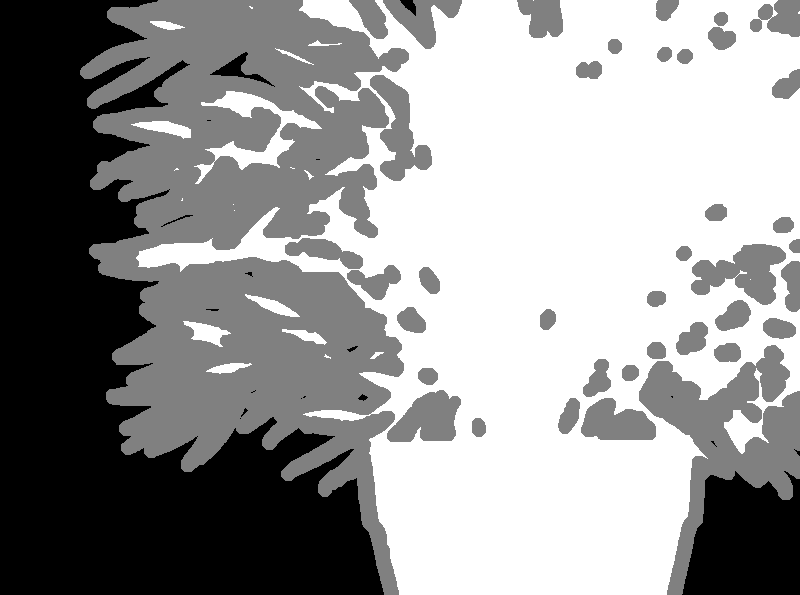}}}}
	\qquad
	\subfloat[Estimated $\alpha$]{{\fbox{\includegraphics[width=3cm]{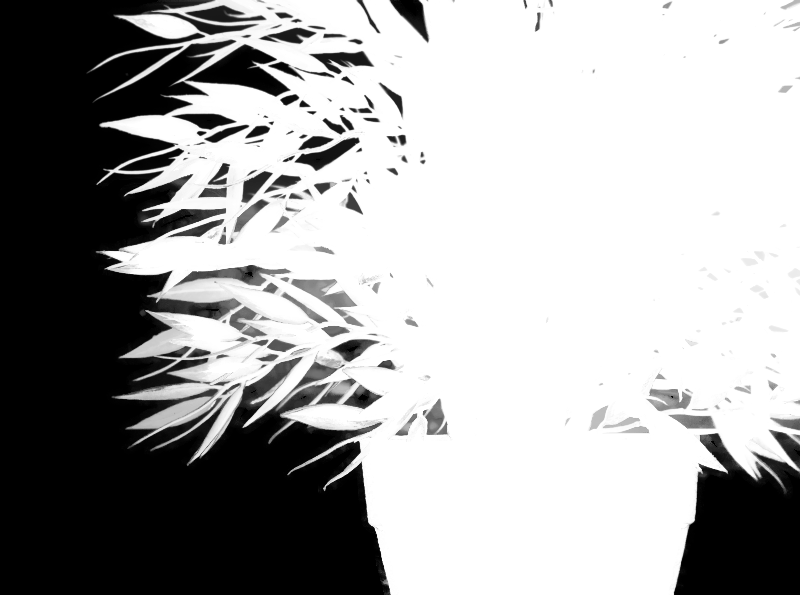}}}}
	\qquad
	\subfloat[Estimated $F$]{{\fbox{\includegraphics[width=3cm]{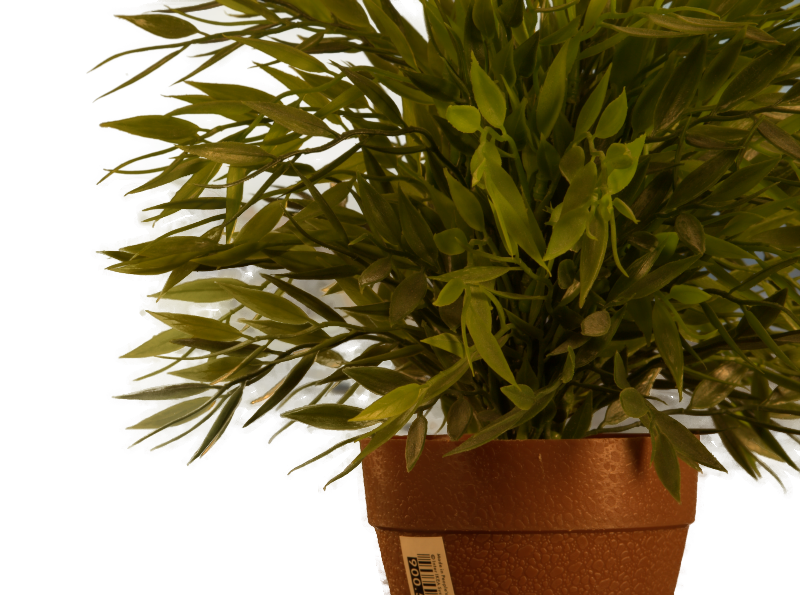}}}}\\
	\caption{Example output of the PyMatting library. Image from \cite{rhemann2009perceptually}.}
\end{figure}
The following code snippet demonstrates the usage of the library:

\definecolor{string}{HTML}{1f77b4}
\definecolor{symbol}{HTML}{ff7f0e}
\definecolor{keyword}{HTML}{d62728}
\definecolor{identifier}{HTML}{000000}

\vspace{2mm}

\texttt{\textcolor{keyword}{from}~\textcolor{identifier}{pymatting}~\textcolor{keyword}{import}~\textcolor{symbol}{\char`\* }}

\texttt{\textcolor{identifier}{image}~\textcolor{symbol}{\char`\= }~\textcolor{identifier}{load\_image}\textcolor{symbol}{\char`\( }\textcolor{string}{\char`\" plant\char`\_ image\char`\. png\char`\" }\textcolor{symbol}{\char`\, }~\textcolor{string}{\char`\" RGB\char`\" }\textcolor{symbol}{\char`\) }}

\texttt{\textcolor{identifier}{trimap}~\textcolor{symbol}{\char`\= }~\textcolor{identifier}{load\_image}\textcolor{symbol}{\char`\( }\textcolor{string}{\char`\" plant\char`\_ trimap\char`\. png\char`\" }\textcolor{symbol}{\char`\, }~\textcolor{string}{\char`\" GRAY\char`\" }\textcolor{symbol}{\char`\) }}

\texttt{\textcolor{identifier}{alpha}~\textcolor{symbol}{\char`\= }~\textcolor{identifier}{estimate\_alpha\_cf}\textcolor{symbol}{\char`\( }\textcolor{identifier}{image}\textcolor{symbol}{\char`\, }~\textcolor{identifier}{trimap}\textcolor{symbol}{\char`\) }}

\texttt{\textcolor{identifier}{foreground}~\textcolor{symbol}{\char`\= }~\textcolor{identifier}{estimate\_foreground\_cf}\textcolor{symbol}{\char`\( }\textcolor{identifier}{image}\textcolor{symbol}{\char`\, }~\textcolor{identifier}{alpha}\textcolor{symbol}{\char`\) }}

\texttt{\textcolor{identifier}{cutout}~\textcolor{symbol}{\char`\= }~\textcolor{identifier}{stack\_images}\textcolor{symbol}{\char`\( }\textcolor{identifier}{foreground}\textcolor{symbol}{\char`\, }~\textcolor{identifier}{alpha}\textcolor{symbol}{\char`\) }}

\texttt{\textcolor{identifier}{save\_image}\textcolor{symbol}{\char`\( }\textcolor{string}{\char`\" results\char`\. png\char`\" }\textcolor{symbol}{\char`\, }~\textcolor{identifier}{cutout}\textcolor{symbol}{\char`\) }}

\vspace{2mm}

The $\texttt{estimate\_alpha\_cf}$ method implements closed-form alpha estimation \citep{levin2007closed}, whereas the  $\texttt{estimate\_foreground\_cf}$ method implements the closed-form foreground estimation \citep{levin2007closed}. 
The method $\texttt{stack\_images}$ can be used to compose the foreground onto a new background.

More code examples at different levels of abstraction can be found in the documentation of the toolbox. PyMatting can be easily installed through pip.

\section{Performance Comparison}
\begin{figure}
	\centering
	\includegraphics[width=15cm]{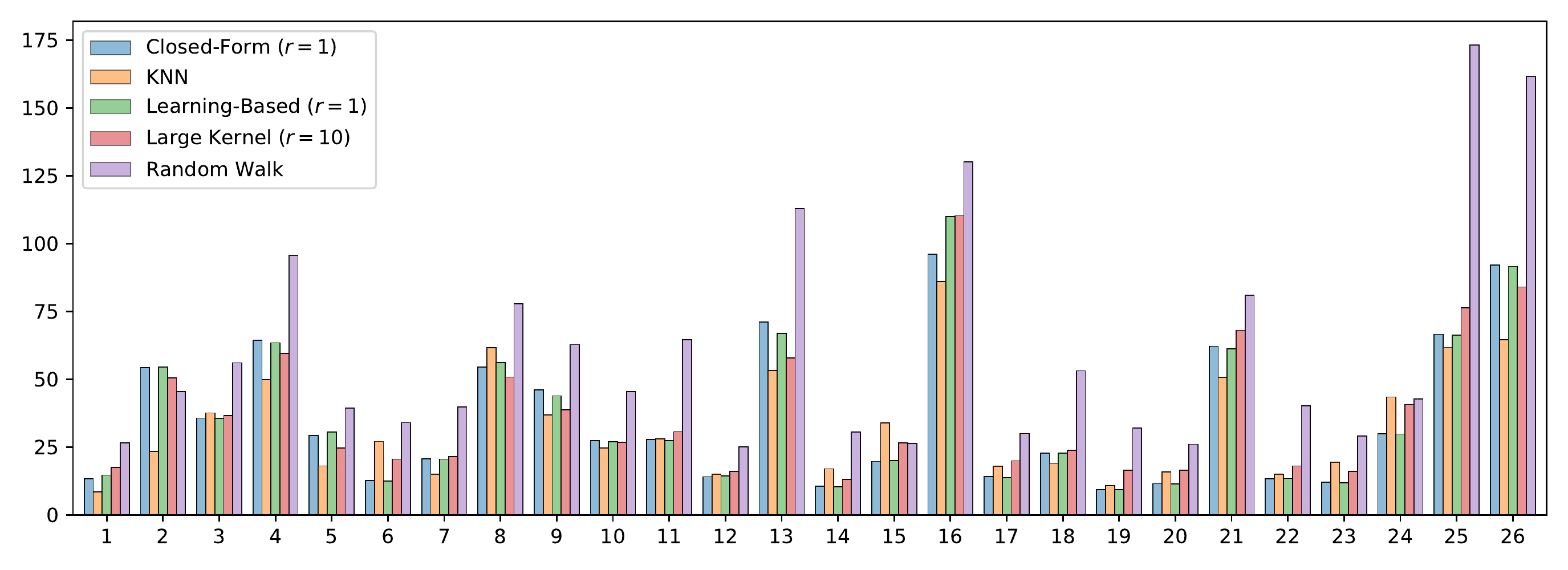}
	\caption{Mean squared error of the estimated alpha matte to the ground truth alpha matte.}
	\label{errors}
\end{figure}
Since all of the considered methods require to solve a large sparse systems of linear equations, an efficient solver is crucial for good performance. 
Therefore, the PyMatting package implements the conjugate gradients method \citep{hestenes1952methods} together with different preconditioners that improve convergence:
Jacobi, \mbox{V-cycle} \citep{lee2014scalable} and thresholded incomplete Cholesky decomposition \citep{jones1995improved}. 

To evaluate the performance of our implementation, we calculate the mean squared error on the unknown pixels of the benchmark images of \citet{rhemann2009perceptually}. 
Figure \ref{errors} shows the  mean squared error to the ground truth alpha matte.
Our results are consistent with the results achieved by the authors implementation (if available).

\begin{figure}[h]
	\centering
	\subfloat[Peak memory usage for each solver in MB]{{\includegraphics[width=7.0cm]{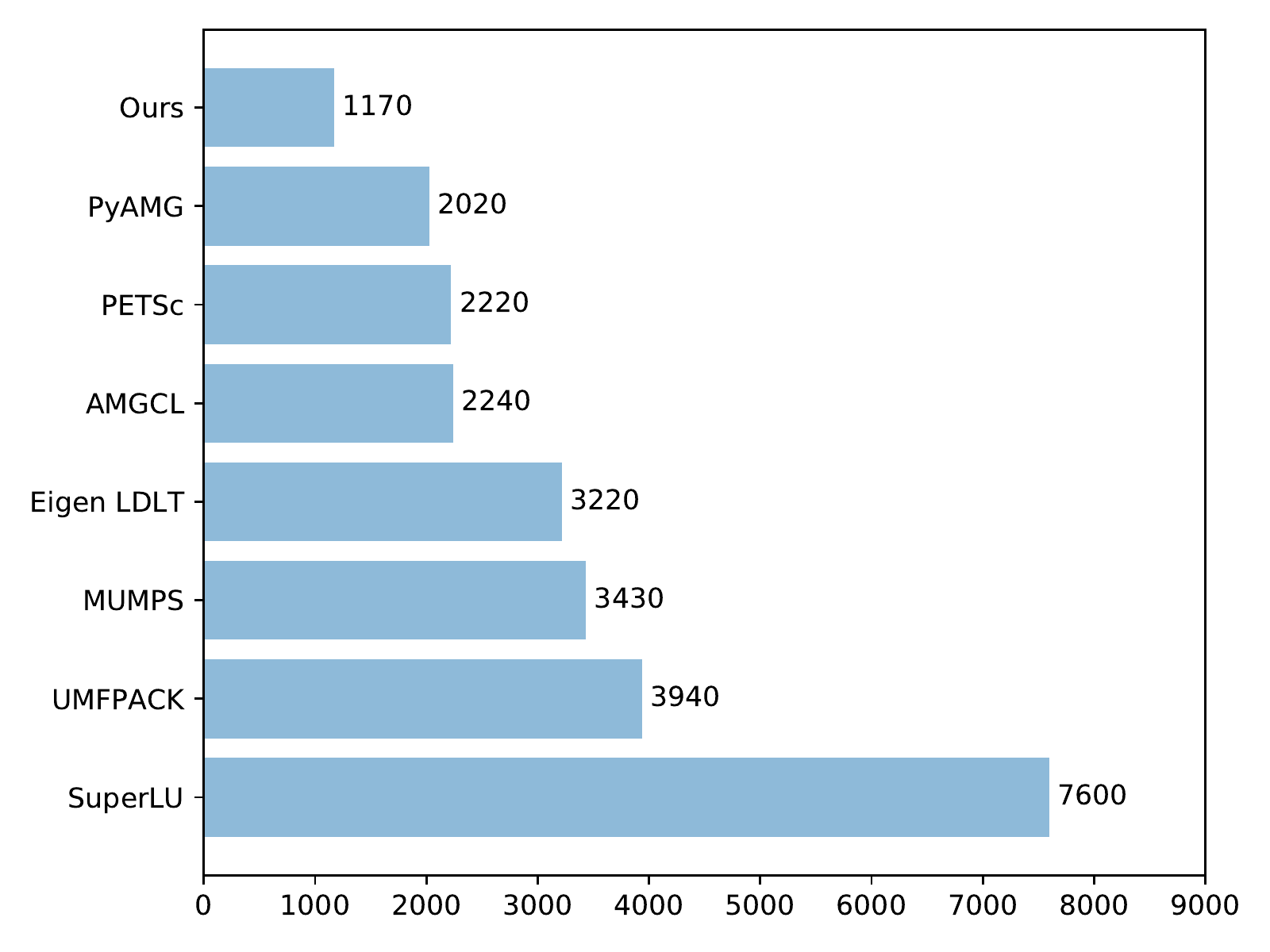} }}
	\qquad
	\subfloat[Mean running time of each solver in seconds]{{\includegraphics[width=7.0cm]{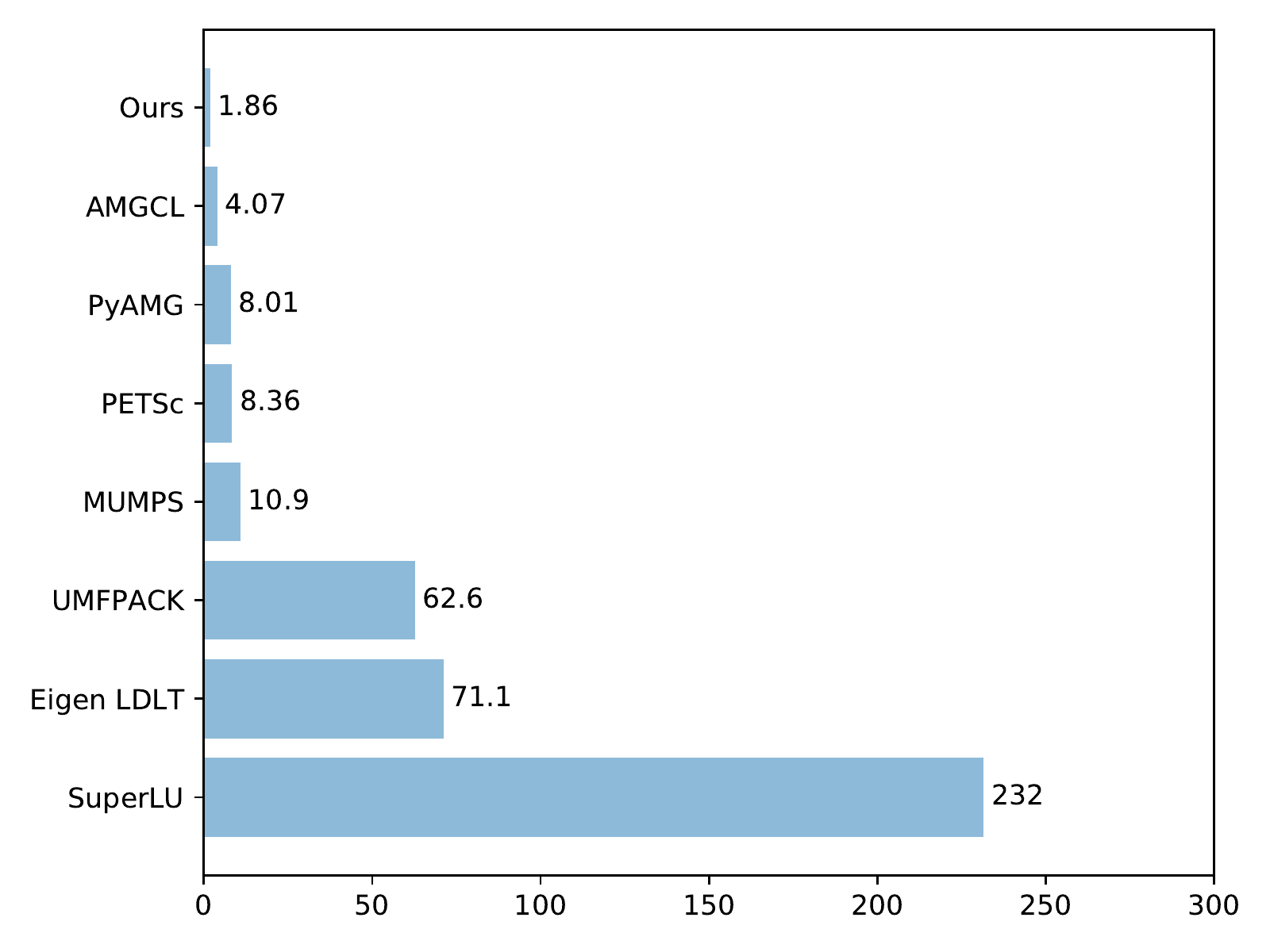} }}
	\caption{Comparison of computational time and peak memory usage of our implementation of the preconditioned CG method with other solvers for closed-form matting.}			\label{memoryusage}
\end{figure}

\begin{figure}
	\centering
	\includegraphics[width=9.0cm]{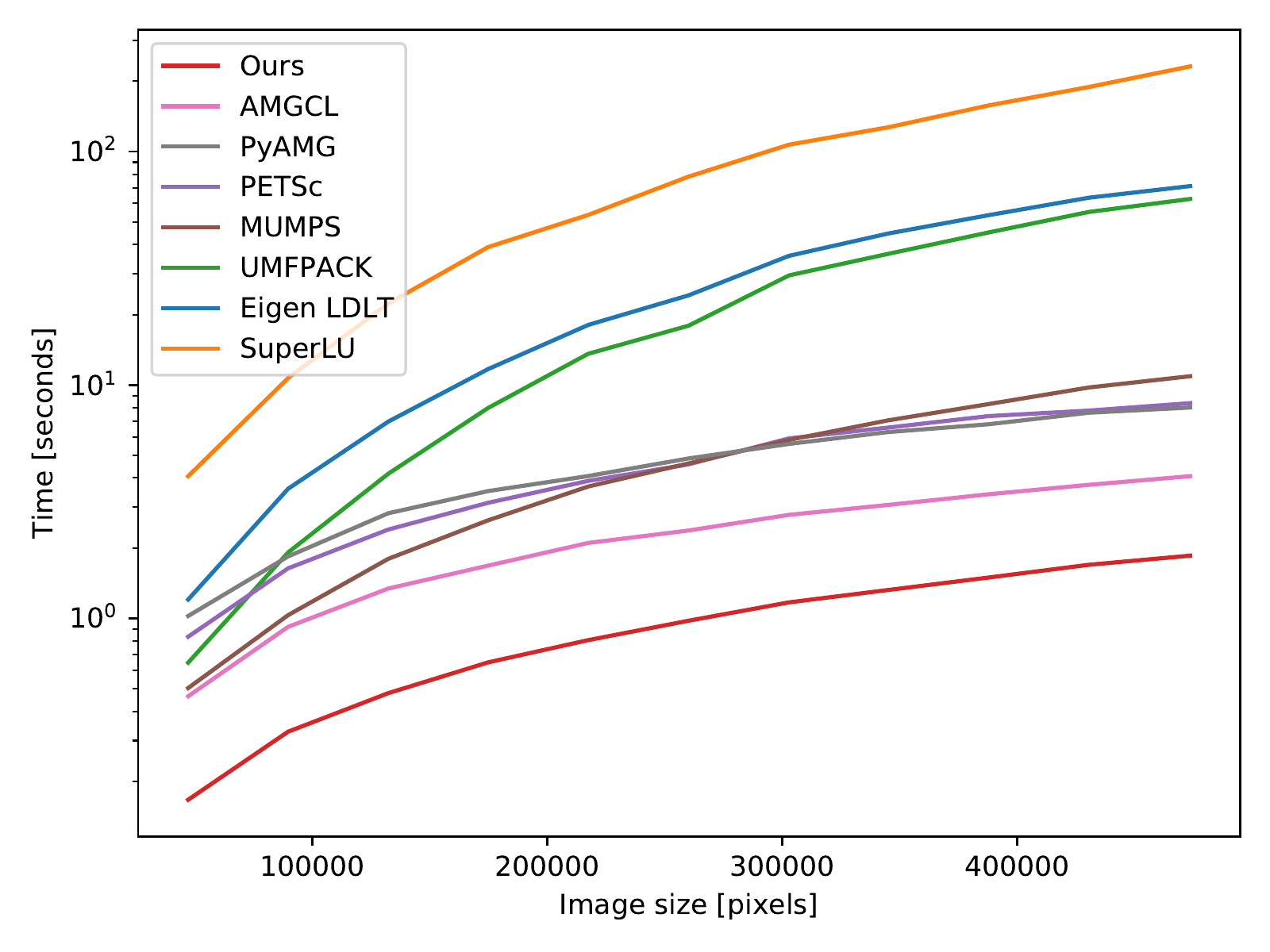}
	\caption{Comparison of runtime for different image sizes.}
	\label{runtimes}
\end{figure}

We compare the computational runtime of our solver with other solvers: PyAMG \citep{pyamg}, UMFPACK \citep{umfpack}, AMGCL \citep{amgcl}, MUMPS \citep{MUMPS-a, MUMPS-b}, Eigen \citep{eigen} and SuperLU \citep{li1999superlu}. Figure \ref{runtimes} shows that our implemented conjugate gradients method in combination with the incomplete Cholesky decomposition preconditioner outperforms the other methods by a large margin. For the iterative solver, we use an absolute tolerance of $10^{-7}$, which we scale with the number of known pixels, i.e. pixels that are either marked as foreground or background in the trimap. The benchmarked linear system arises from the matting Laplacian by \citet{levin2007closed}. Figure \ref{memoryusage} shows that our solver also outperforms the other solvers in terms of memory usage.

\section{Compatibility and Extendability}
The PyMatting package has been tested on Windows 10, Ubuntu 16.04 and macOS 10.15.2.
The package can be easily extended by adding new definitions of the graph Laplacian matrix. 
We plan on continuously extending our toolbox with new methods.

\bibliographystyle{plainnat}
\bibliography{pymatting}

\end{document}